\documentclass[sigconf,authorversion,nonacm]{acmart}

\usepackage{booktabs} 
\usepackage{multirow}

\setcopyright{none}

\begin{document}
\title{Collective Anomaly Perception During Multi-Robot Patrol: Constrained Interactions Can Promote Accurate Consensus}

\renewcommand{\shorttitle}{Collective Anomaly Perception During Multi-Robot Patrol}

\author{Zachary R. Madin}
\orcid{0009-0005-0310-4504}
\affiliation{%
  \institution{University of Bristol}
  \city{Bristol}
  \country{United Kingdom}
}
\email{zachary.madin@bristol.ac.uk}

\author{Jonathan Lawry}
\affiliation{%
  \institution{University of Bristol}
  \city{Bristol}
  \country{United Kingdom}
}
\email{j.lawry@bristol.ac.uk}

\author{Edmund R. Hunt}
\affiliation{
  \institution{University of Bristol}
  \city{Bristol}
  \country{United Kingdom}
  }
\email{edmund.hunt@bristol.ac.uk}

\renewcommand{\shortauthors}{Madin et al.}

\begin{abstract}
An important real-world application of multi-robot systems is multi-robot patrolling (MRP), where robots must carry out the activity of going through an area at regular intervals. Motivations for MRP include the detection of anomalies that may represent security threats. While MRP algorithms show some maturity in development, a key potential advantage has been unexamined: the ability to exploit collective perception of detected anomalies to prioritize the location ordering of security checks.  This is because noisy individual-level detection of an anomaly may be compensated for by group-level consensus formation regarding whether an anomaly is likely to be truly present. Here, we examine the performance of unmodified idleness-based patrolling algorithms when given the additional objective of reaching an environmental perception consensus via local pairwise communication and a quorum threshold. We find that generally, MRP algorithms that promote physical mixing of robots, as measured by a higher connectivity of their emergent communication network, reach consensus more quickly. However, when there is noise present in anomaly detection, a more moderate (constrained) level of connectivity is preferable because it reduces the spread of false positive detections, as measured by a group-level F-score. These findings can inform user choice of MRP algorithm and future algorithm development.
\end{abstract}

\maketitle

\section{Introduction}
An important application for multi-robot systems (MRS) is multi-robot patrol (MRP), where multiple robots are required to somehow coordinate their behaviour such that all points in the environment requiring surveillance are visited regularly. Several multi-robot patrol algorithms have been developed over the past 20 years or so, which commonly aim to minimize the maximum and average `idleness' \cite{machado_multi-agent_2003} (return times) on each vertex in a patrol graph, using a variety of centralized and distributed robot coordination methods \cite{camarinha-matos_survey_2011, huang_survey_2019}. Depending on the real-world deployment context, one aim of such patrols might be to detect subtle anomalies representing a potential security threat, for example the electromagnetic activity of an eavesdropping device planted by an adversary. Such anomalies, when detected, are quite likely to be false positives, i.e. to be false alarms caused by noise in sensors or in the environment. An important opportunity for MRS, then, is to help prioritize limited security resources, by advising a user on system-level consensus on an anomaly's presence. In the course of MRP, one or more robots can make more than one inspection of an anomaly to obtain more information, and/or the detection system can benefit from the mechanisms of consensus formation to slow the spread of incorrect opinions (`misinformation'). Additional anomaly inspections may be deliberate (i.e. planned by the robots following an initial detection) or occur spontaneously in the course of ongoing patrol. In an adversarial context, it would be important not to unduly compromise idleness minimization to make additional inspections, as this could be a source of vulnerability in a multi-stage, multi-location attack \cite{sless_lin_multi-robot_2019}. In the first instance, then, one may wish to retain the primary objective of idleness minimization, while allowing a consensus formation process regarding anomaly location to take place during the course of normal, longer-term operations. Maintaining typical patrolling behaviours may also be less likely to alert an adversary to their discovery. 

In this work, we use a realistic ROS-based MRP simulator to examine the performance of unmodified patrolling algorithms when given the secondary objective of reaching a perceptual consensus on anomaly presence, via local
pairwise communication and a quorum decision threshold. We consider the accuracy of the system-level perception in relation to the level of noise present in the anomaly detection process. Overall, MRP algorithms that promote more physical mixing of robots have faster consensus, given the assumption of local communications. However, when anomaly detection noise increases, a more moderate communication connectivity is preferable to suppress false positives. These findings could help to inform user choice of patrolling algorithm, given expectations around detection noise and attack frequency. We provide more background on relevant research in Section \ref{sec:background}, and detail our methodology in terms of simulator, collective decision-making mechanisms, and graph metrics in Section \ref{sec:experiment}. In Section \ref{sec:results} we present simulation results and in Section \ref{sec:discussion} we provide some conclusions and plans for future work. 

\section{BACKGROUND}\label{sec:background}

We briefly review MRP algorithms and collective decision-making and perception. We also consider a graph-based metric to describe the emergent communication network between patrolling robots with local belief exchange.  

\subsection{Multi-Robot Patrolling Algorithms}
\begin{figure}
  \centering
  \includegraphics[width=\columnwidth]{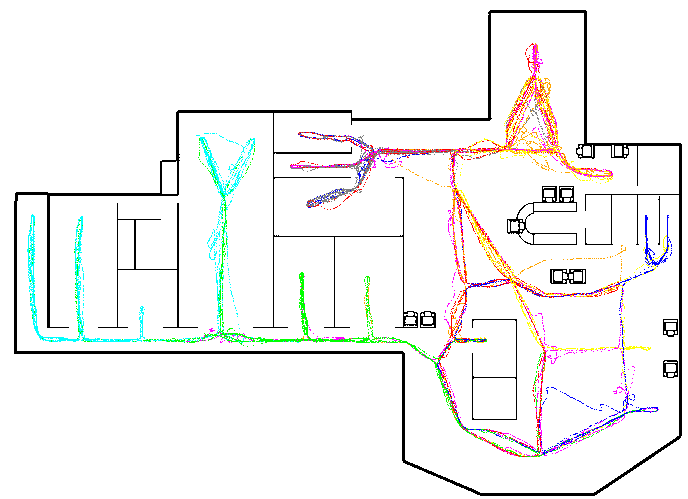}
  \caption{Trajectory plot of a system of eight robots executing the SEBS patrol algorithm for over an hour on the `Cumberland' map , with each unique color representing a single robot and its patrol route.}
  \label{fig:agent_footprint}
\end{figure}

MRP generally focuses on a static environment to be patrolled. This is defined as an area of interest that has nodes that are to be visited, in order to minimize idleness of each node. The term \textit{instantaneous node idleness} was proposed by Machado et al. \cite{machado_multi-agent_2003} and is used extensively in the MRP field. The application of this metric is to define the number of cycles or time since a node was last visited. In addition to this term there is also \textit{`instantaneous graph idleness'} which is the average idleness of each node in the entire region or graph. Given a map or graph to patrol by an MRS (e.g. Figure~\ref{fig:agent_footprint}), one fundamental problem to solve is the apportionment of locations (nodes) to visit by different robots, where efficient visitation will physically spread the agents between nodes. One approach is an auction-based system where each robot submits a bid either to a central auctioneer \cite{hwang_cooperative_2009} or between robots in a local proximity \cite{choi_consensus-based_2009}. Much work has been done to take developments from Game Theory and apply it to the robot patrol problem. Owing to the malicious nature of an attacker in a system, different patrolling strategy approaches are required \cite{trejo_stackelberg_2015}. With predictability comes weakness to attackers or intruders who may observe the pattern of patrol and gain access during robot downtime \cite{basilico_patrolling_2012}. A more realistic algorithmic performance benchmark, beyond idleness minimization, needs to take such predictability into account \cite{Ward2023}. Other recent approaches have taken advantage of machine learning capabilities in order to derive a `learned' patrol strategy to offer solutions to adversarial patrolling as traditional methods are subject to exploitation by attackers \cite{bogert_multi-robot_2018, zhou_bayesian_2019}. These reinforcement methods offer benefits when it comes to managing potential attackers, but often do not generalize well to different environments \cite{liu_realsimreal_2020}. An in-depth study of the latest approaches to the problem of multi-robot patrolling and their respective benefits and shortcomings can be found in \cite{basilico_recent_2022}. 

\begin{table*}

\centering
\Large


\begin{tabular}{lcccl}\toprule
\large 
Short name & \large Full name                                    & \large Patrol Strategy & \large Decentralized? &  \\ \cline{1-4}
\normalsize CBLS\cite{portugal_cooperative_2016}       &  \normalsize Concurrent Bayesian Learning Strategy           & \normalsize Bayesian Learning  & \normalsize Yes            &  \\
\normalsize CGG\cite{portugal_msp_2010}        & \normalsize Cyclic Algorithm for Generic Graphs                      & \normalsize Hamiltonian Path          & \normalsize No             &  \\
\normalsize CR\cite{machado_patrulha_2002}         & \normalsize Conscientious Reactive                               & \normalsize Reactive          & \normalsize Yes            &  \\
\normalsize DTAG\cite{farinelli_distributed_2017}       & \normalsize Dynamic Task Assignment Greedy                  & \normalsize Utility Function  & \normalsize Yes            &  \\
\normalsize DTAP\cite{farinelli_distributed_2017}       & \normalsize Dynamic Task Assignment Auction                 & \normalsize Utility Auctioneer  & \normalsize Yes            &  \\
\normalsize GBS\cite{portugal_distributed_2013}        & \normalsize Greedy Bayesian Strategy                         & \normalsize Bayesian      & \normalsize No             &  \\
\normalsize HCR\cite{almeida_a_patrulhamento_2003}        & \normalsize Heuristic Conscientious Reactive              & \normalsize Heuristic Reactive          & \normalsize Yes            &  \\
\normalsize HPCC\cite{almeida_a_patrulhamento_2003}       & \normalsize  Heuristic Pathfinder Conscientious Cognitive & \normalsize Heuristic Pathfinder          & \normalsize Yes            &  \\
\normalsize RAND\cite{machado_multi-agent_2003}       & \normalsize Random                                            & \normalsize Random Selection          & \normalsize Yes            &  \\
\normalsize SEBS\cite{portugal_distributed_2013}       & \normalsize State Exchange Bayesian Strategy                 & \normalsize State \& Bayes Exchange         & \normalsize Yes            & \\\bottomrule

\end{tabular}

\caption{MRP algorithms examined in this study}
\label{tab:alg_behaviour}
\end{table*}

A broad selection of ten existing multi-robot patrolling algorithms from the literature, designed to minimize idleness, are used here without modification. These algorithms are structured in different ways and as such, patrolling robots exhibit a variety of behavior (Table~\ref{tab:alg_behaviour}). More detail on the structure and operation of the algorithms is presented in \cite{camarinha-matos_survey_2011, portugal_decision_2012, machado_multi-agent_2003}. As we go on to discuss, the algorithms vary considerably in the extent to which robots become spatially mixed, which is significant when they have local communication. 

\subsection{Collective Decision-Making and Perception}

One approach to managing the large quantity of information that is gathered in a multi-robot system is for each robot to contribute its  `opinions' or `beliefs' about the state of the world, and according to some global or local mechanism, allow the group to determine which to collectively adopt, referred to as reaching a \textit{consensus}. Often owing to communication constraints, and also to prevent rapid noise propagation, consensus formation can be performed at a local level, between individual robots. How the consensus between all robots is formed depends on the belief exchange operator that is used to respond to information. One approach is to use voting model methods, where input from each robot is submitted into a voting system in order to determine the opinion of the group \cite{hassanzadeh_building_2013, shan_discrete_2021}, but this approach can often suffer from noisy measurements from individual robots \cite{crosscombe_robust_2017}. An approach that seeks to solve the issue caused by noisy measurements was initially proposed by Perron et al. \cite{perron_using_2009}, whereby a third truth state of `\textit{uncertain}' is employed as a compromise between certain of false or certain of negative. This was applied directly to multi-agent systems in \cite{crosscombe_exploiting_2016}, which showed that the application of this Kleene uncertain ternary logic can result in better performance than a standard Boolean truth pair in noisy environments.

In the collective learning literature, a common task is for agents to perceive a feature of the environment and come to a consensus on the state of that feature. This has been abstracted as a perception of white to black floor tiles (e.g. \cite{shan_discrete_2021}), with the ratio being the element being voted on. This scenario shares similarities with the patrolling problem in regard to communication and data exchange \cite{valentini_best--n_2017}, but the information is of a single dimension or value -- the ratio of black to white. In the patrolling problem there is a single value for each area of interest, with all nodes having the same importance as another.

With the collective learning and communication of information, comes the ability for erroneous information to propagate throughout the system and cause a false consensus to be reached. Often in the literature an assumption is made that the world is static and fixed -- that there is no change in the real values of the measured world. In a dynamic environment -- such as a realistic patrolling problem -- there are changes in the world that need to be recorded, such as the activity of an attacker, or variability in the physical environment affecting sensor readings (`noise'). Some recent work has shown that robots adapt better in environments that are subject to change when they have a constrained communication, preventing preemptive consensus formation \cite{talamali_when_2021}.

\subsection{Communication Graph Connectivity}

Robots in a multi-robot system can be dispersed over a wide area, and the ability for robots to maintain communication with one another is dependent staying within local communication range. If a robot is isolated or weakly connected to the communication graph, it will be unable to exchange information that it gathers or receive new information from other robots. It is therefore useful to use a measure of connectivity that adequately describes the interconnections between the robots and the strength of said connections.
One of the most extensively used metrics to measure this in the literature is the \textit{algebraic connectivity} or `Fiedler value' \cite{fiedler_algebraic_1973}. Mathematically, it is the second smallest eigenvalue of the Laplacian matrix for the graph, where the Laplacian matrix is the degree matrix subtracted from the adjacency matrix. Conceptually, it is a measure of connections between nodes as well as the strength of the edges between nodes within a graph.
The algebraic connectivity has a wide range of applications in determining the strength of connection of nodes within a system, most notably in wireless communications where adequate coverage and interconnections are highly desirable properties \cite{zavlanos_distributed_2008, lin_composable_2008, xue_number_2004}. The measure of algebraic connectivity can be highly correlated with performance under certain circumstances such that it can be used as an input when designing the control algorithms for mobile robots. Zavlanos and Pappas used the algebraic connectivity of a team of mobile robots' communications as a way of controlling their spatial distribution \cite{zavlanos_graph-theoretic_2011}. Other work has been done to use algebraic connectivity as part of the feedback input to a system within applications such as formation control \cite{poonawala_collision-free_2015}. Here, we use it to assess the emergent communication networks resultant from robots moving according to various MRP algorithms.

\begin{figure}
  \centering
  \includegraphics[width=\columnwidth]{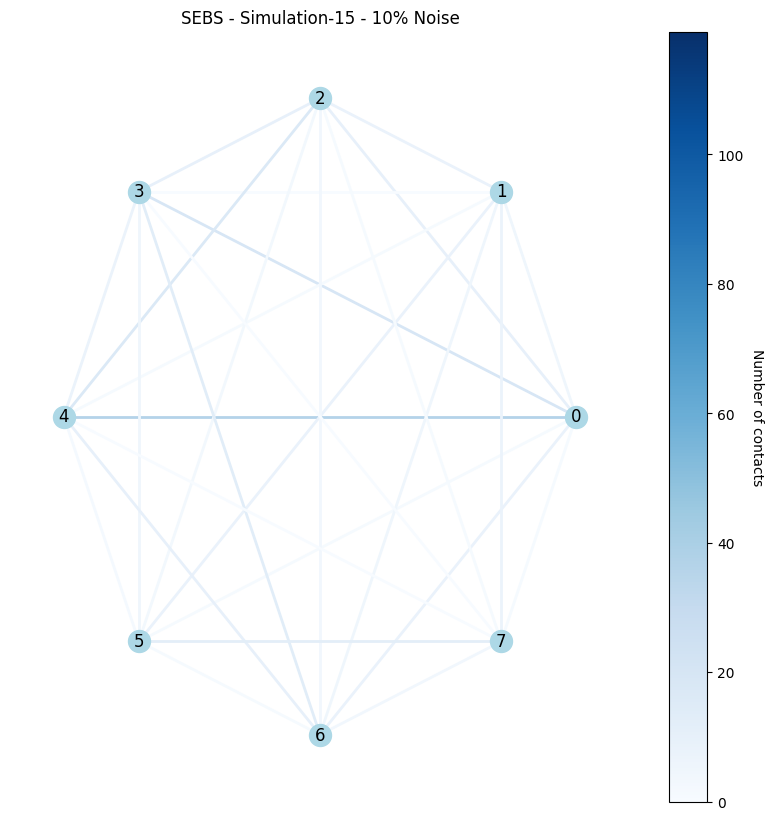}
  \caption{Example social network of eight robots executing SEBS algorithm on `Cumberland' map for 1 hour. Nodes are individual robots, with edges representing the number of communications between respective robots. Darker shade indicates more frequent communication between robots.}
  \label{fig:social_net}
\end{figure}

\section{Experimental Set Up}\label{sec:experiment}

We examine the performance of a multi-robot system (MRS) undertaking multi-robot patrol (MRP) when there is one, and only one, true anomaly present. The anomaly is always located in the same place on the patrol graph (`node 30'). When there is detection noise, false positive anomaly detections occur, and we examine the ability of the collective opinion dynamics to suppress these. 

\subsection{Simulator}

The simulation environment used in this study is ROS Patrolling Sim\footnote{https://wiki.ros.org/patrolling\_sim}, which is described in \cite{portugal_effective_2013}. The simulator is built using the Stage\footnote{https://github.com/rtv/Stage}  simulation package \cite{gerkey_playerstage_2003} and integrated into the ROS (Robot Operating System) framework, which enables easy transfer to real robotic systems.

Each robot in the simulation is modeled as a differential drive robot, with a laser rangefinder and odometry from wheel encoders with a drift error model. 
As the modality of the actual sensor is not the focus of this paper, anomalies to be detected are modeled as measurements that are made with a generic sensor. It is assumed that each observation of an anomaly is instantaneous upon reaching a patrol graph node, and subject to some noise probability value according to the experiment. Upon arrival to a node on the graph, the robot searches for an anomaly and records a true or false depending on world state and noise model for that node. Given a probability value of 5\%, there is a 5\% chance that the sensor records the opposite value that is present in the world. Upon revisiting that same node, there is a new chance of making an incorrect measurement.
Robots are modeled as having an omni-directional antenna that is capable of communicating in a defined radius with other robots in the vicinity, regardless of obstacles or walls that may be between each robot. Communication is modeled as a one-to-one (pairwise) exchange, with a minimum timeout/delay period before repeated communications between the same robots. Upon initializing communication with another robot, a pairwise comparison of belief states occurs resulting in both robots having identical beliefs.

\begin{figure*}
  \centering
  \includegraphics[width=0.7\textwidth]{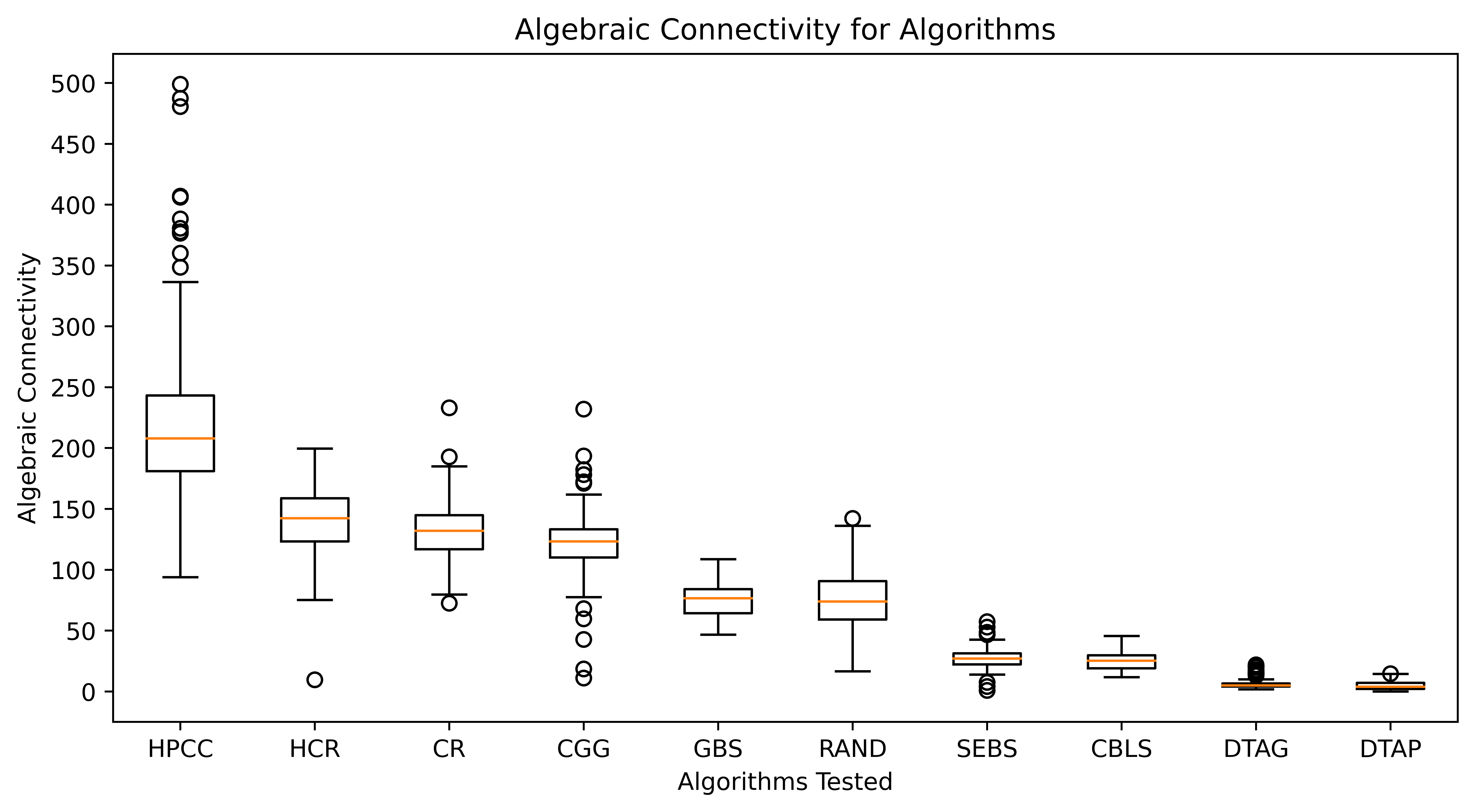}
  \caption{Algebraic Connectivity results for each algorithm tested }
  \label{fig:alg_conn_boxplot}
\end{figure*}

\begin{figure*}

        \includegraphics[width=\textwidth]{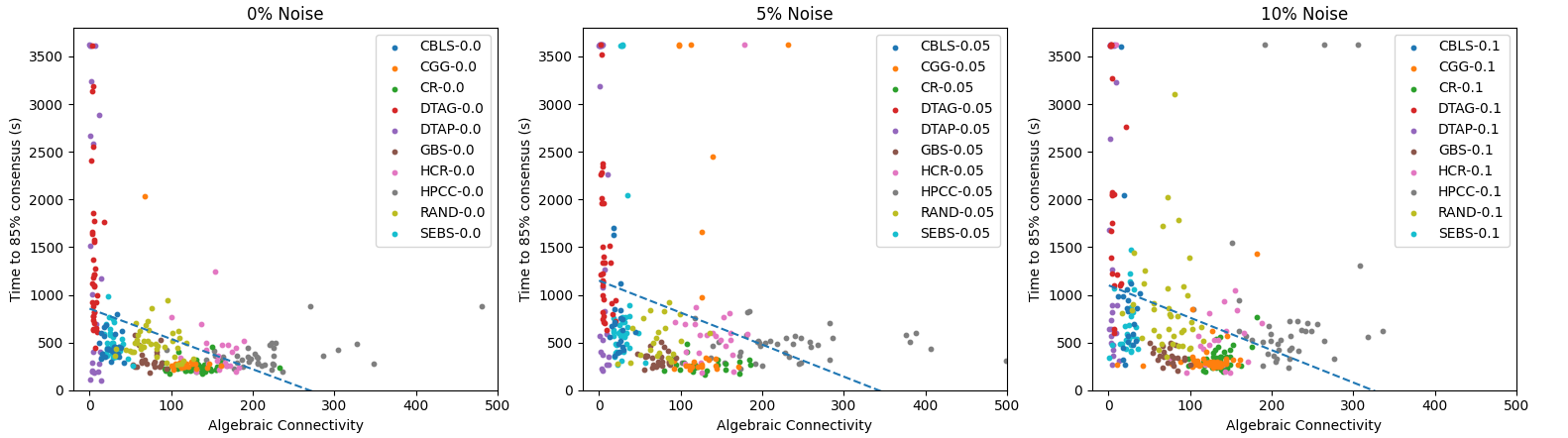}
        
    \caption{Time to reach 85\% consensus on accurate view of world state against algebraic connectivity for each algorithm and noise level.}
    \label{fig:consensus_alg_conn}
\end{figure*}

\begin{figure*}
    \centering
        \includegraphics[width=\textwidth]{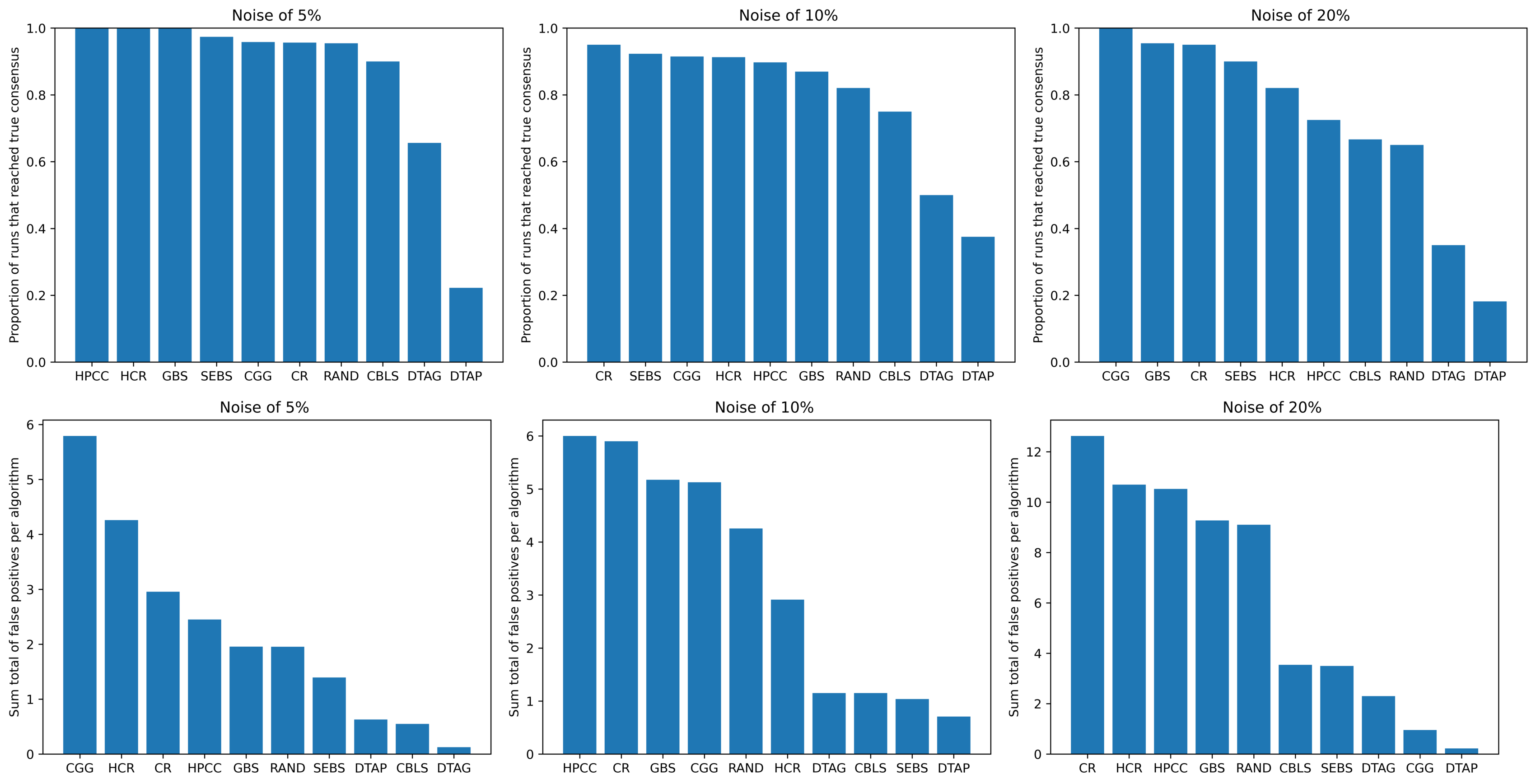}
    \caption{For 20 simulation runs, Top: Proportion of runs that reached consensus ($> 85\%$ or at least 7 of 8 robots) on particular node containing true positive (real security threat); Bottom: Average of count false positive consensus on other nodes.}
    \label{fig:Tp_Fp_sum}
\end{figure*}

\subsection{Opinion Dynamics}

In a world with $n$ nodes on the patrol graph, robots have a set of belief values that can be either $1, 0$ or $\frac{1}{2}$, meaning true, false and uncertain respectively. Each robot maintains a set of measured belief states of the world according to the number of nodes that exist in the map $M = \{ m_0, \cdots , m_n \}$, where $m_i$ is the robot's value for node $i$. When two robots compare beliefs, they update their belief values according the fusion truth table shown in Table~\ref{tab:fusion}.

\setlength{\tabcolsep}{10pt} 
\renewcommand{\arraystretch}{1.5} 
\begin{table}[H]
\centering
\begin{tabular}{cc|ccc}

\multicolumn{1}{l}{}      &             & \multicolumn{3}{c}{$B(p_i)$}{}            \\

  & $\bigodot$  & 0           & $\frac{1}{2}$ & 1           \\

\cmidrule{2-5}
 & 0           & 0           & 0           & $\frac{1}{2}$ \\
\multirow{1}{*}{$B'(p_i)$}                          & $\frac{1}{2}$ & 0           & $\frac{1}{2}$ & 1           \\
                          & 1           & $\frac{1}{2}$ & 1           & 1          
\end{tabular}
\caption{Belief fusion truth table for beliefs $B$ and $B'$}
\label{tab:fusion}
\end{table}

Using this belief fusion provides benefits to the system as a whole, as each robot can gain information on regions of the world that they have not explored. 
As the simulation world is subject to anomaly detection noise at individual nodes, the ability to reject noise at the system level is highly desirable. When two robots that begin the communication process have conflicting belief states on the same node, their resultant value is that they are both uncertain about the value of that node. This ability to reconcile conflicting information also extends to measuring a node directly. When an agent revisits a node and takes a measurement which does not match its prior belief, the belief fusion in Table~\ref{tab:fusion} is performed between the prior and subsequent beliefs.
By employing the use of pairwise communication, the system as a whole is able to collectively learn the state of the world faster than a single robot. The overall benefit of limiting the information breadth by using this ternary belief state model over, say, a voting model is that the three-value system is better suited to reject noisy measurements or information from faulty agents \cite{crosscombe_robust_2017}.

\subsection{Error and F-Score}

In order to judge the performance of the system at arriving at an accurate consensus of the world state, a metric which will be referred to as \textit{system error} is used, where the error is defined as:
\begin{equation}
    E = \frac{1}{m}\frac{1}{n}\sum_{j=0}^{m}\sum_{i=0}^{n} \left| B_j (p_i) - S^*(p_i) \right|
    \label{eq:accuracy}
\end{equation}
Where \textit{m} is the number of nodes on the graph, \textit{n} is the number of robots in the experiment, \textit{B} is the belief for a node, \textit{S$^*$} is the true world state and \textit{p} is the agent's belief at a given node
\cite{crosscombe_impact_2022}.

Another system-level performance metric used is an F-score, which is introduced here in Equation~\ref{eq:f-score}. This represents the harmonic mean of the precision and recall of the system, which gives a way to compare the recorded results with the real world in order to determine the effect of noise on the performance of each algorithm. Terms in Equation~\ref{eq:f-score} are $t_p$, the count of true positives, $t_n$, true negatives, $f_p$, false positives, $f_n$, false negatives and $u$, the number of unmeasured nodes. These are given a half weighting as per the uncertain belief in Table~\ref{tab:fusion}, to penalize the performance metric if nodes remain unknown. The use of this system-level F-score provides a useful insight into the performance of the system with regard to precision of the data that is collected. This scoring will be given for each algorithm at the end of the experiment when an expected consensus is to be reached.

\begin{equation}
    F = \frac{2 (t_p + t_n)}{2(t_p + t_n) + f_p + f_n + \frac{u}{2}}
    \label{eq:f-score}
\end{equation}

\subsection{Experiment Parameters}
For each experiment run, each robot is initialized at the same position regardless of patrol algorithm to ensure comparability. Each set of experiments are run on the same default map (`Cumberland') that has 40 patrol graph nodes (Figure~\ref{fig:agent_footprint}). This map is representative of a realistic building environment. The simulations have a system of eight robots, a group size that is both realistic and reduces the number of interference or collision events that can occur at higher robot densities \cite{portugal_decision_2012}. There were 20 simulation runs per algorithm. Table~\ref{tab:exp_params} shows some further simulation parameters. 

\begin{table}[H]
\begin{tabular}{lc}\toprule
\textbf{Experiment Parameter} & \textbf{Value}\\\midrule
Robot communication range                            & 5 m                   \\
Communication timeout period                           & 30 s                  \\
Number of robots                               & 8                    \\
Time of experiment                             & 3600 s                \\
Measurement noise probability                  & 0, 5, 10 \& 20\%     \\
Simulations per algorithm \& noise level            & 20                   \\\bottomrule

\end{tabular}

\caption{Simulated experiment parameter values}
\label{tab:exp_params}
\end{table}

In order to compute the simulation runs, the software packages were `containerized' and executed in parallel on a cloud-based compute service. The cloud service utilizes a 22-core Intel Xeon Gold 6238 with 16GB of RAM.

\section{Simulation Results and Discussion}\label{sec:results}

A boxplot of the algebraic connectivity for each algorithm is presented in Figure~\ref{fig:alg_conn_boxplot}. This shows a wide variety in communication intensity between robots, with the HPCC, HCR, CR and CGG algorithms leading to noticeably higher algebraic connectivity than the other six algorithms tested. At the other end of the scale, DTAP and DTAG have very low connectivity, while GBS, RAND, SEBS and CBLS have intermediate connectivity. Figure~\ref{fig:alg_conn_boxplot} seems to suggest, then, three approximate groupings of algorithms of high, intermediate and low connectivity. 

There was a negative correlation between a system's algebraic connectivity and the convergence time to 85\% consensus on an accurate world view, i.e. at least 7 of 8 robots completely matching the true state of the world in their beliefs (Figure~\ref{fig:consensus_alg_conn}). There was a Pearson correlation coefficient of $-0.377$ ($p\ll0.001$) for 0\% noise, $-0.313$ ($p\ll0.001$) for 5\% noise, and $-0.299$ ($p\ll0.001$) for 10\% noise. On this figure 20\% noise level is omitted as a majority of the simulations did not reach accurate consensus within the experiment run time. While faster convergence time is welcome if it is to a true positive, it may also have its disadvantages if it is at the expense of larger numbers of false positives. 

Figure~\ref{fig:Tp_Fp_sum} shows that there are several algorithms that reliably give a consensus for a true positive (i.e. a correctly detected anomaly) on the correct node for different noise levels, producing a consensus 100\% of the time, or at least on over 90\% of occasions. This is valuable information that could help to prioritize a security professional's inspections after time away from the building. However, in some cases there were also large numbers of false positives produced elsewhere on the patrol graph (Figure~\ref{fig:Tp_Fp_sum} bottom), averaging around 6 per run in the worst case (CGG) down much less than 1 per run with DTAG, with 5\% noise. As detection noise increases, as anticipated, the number of false positives also increases, and thus it becomes apparent that it might be preferable to employ an algorithm that has good, but not the best, true positive detection performance, in return for cutting down false positives. SEBS and CBLS are notable examples of this, with much lower false positives than some alternatives that perform well in true positives, such as CGG. This trade-off is captured by the F-score.

Table~\ref{tab:results-conv-fscore} shows both the average idleness and F-scores for different noise levels, across the different algorithms tested. Standard deviations $\sigma$ for the results are also shown. The best two algorithms are shown for each metric: very good performance of idleness minimization is found for CBLS and SEBS. For 0\% noise, where false positives do not occur, HPCC and CR obtain the best F-scores, although SEBS is a very close third. As noise increases, RAND and CR lead at 5\% noise, while at 10\% HCR and SEBS lead, and then at 20\% CGG and SEBS are the best performers on F-score. Overall, MRP algorithms with intermediate algebraic connectivity, like SEBS and CBLS, appear to be the best all-rounders.

\begin{table*}

\centering
\resizebox{\textwidth}{!}{%
\Large
\begin{tabular}{lcccccccccc}\toprule

\multicolumn{1}{c}{}          &                      & \multicolumn{1}{c}{}          & \multicolumn{8}{c}{F-Score}                         \\ \cline{4-11} 
\multicolumn{1}{c}{Algorithm} & Avg. Graph Idleness (s)            & \multicolumn{1}{c}{$\sigma$} & \multicolumn{2}{c}{0\% Noise}          & \multicolumn{2}{c}{5\% Noise}          & \multicolumn{2}{c}{10\% Noise}         & \multicolumn{2}{c}{20\% Noise} \\\cmidrule(lr){4-5}\cmidrule(lr){6-7}\cmidrule(lr){8-9}\cmidrule(lr){10-11}
& \multicolumn{1}{c}{} & \multicolumn{1}{c}{}          & Avg. & \multicolumn{1}{c}{$\sigma$} & Avg. & \multicolumn{1}{c}{$\sigma$} & Avg. & \multicolumn{1}{c}{$\sigma$} & Avg.        & $\sigma$       \\\midrule
CBLS & \textbf{59.2}        & 7.7 &  0.962             &    0.0145  &  0.949            &  0.0124                          & 0.916     & 0.0265&   0.871          & 0.0210            \\
CGG  & 100.4                & 17.7&   0.954            &    0.0323  &   0.928           &   0.0462                         &  0.931    & 0.0217&  \textbf{0.962 }          & 0.0257            \\
CR   & 100.3                & 18.0&   \textbf{0.974}   &    0.0008  &  \textbf{0.952}   &  0.0167                          & 0.923      & 0.0181& 0.860            & 0.0314            \\
DTAG & 68.8                 & 6.2 &   0.914            &    0.0646  &   0.910           &   0.0828                         & 0.883     & 0.0595&  0.818           &  0.0633           \\
DTAP & 81.1                 & 21.4&  0.867             &    0.1291  &  0.839            &   0.1312                         &  0.880    & 0.0764 &  0.852           &   0.1057          \\
GBS  & 81.4                 & 16.3&   0.962            &    0.0000  &   0.951           &    0.0122                        &  0.924    & 0.0251& 0.875            &  0.0391           \\
HCR  & 115.3                & 23.5&  0.956             &    0.0200  &   0.933           &    0.0275                        & \textbf{0.944}     & 0.0335&  0.872           & 0.0372            \\
HPCC & 169.2                & 59.7&   \textbf{0.974}   &    0.0000  &   0.951           &   0.0165                         &0.922      &0.0361 & 0.881            &  0.0339           \\
RAND & 99.3                 & 20.9&   0.954            &    0.0025  &  \textbf{0.953}   &   0.0129                         & 0.918     & 0.0260& 0.868            &  0.0314           \\
SEBS & \textbf{60.3}                 & 7.2&   0.970    &    0.0142  &   0.935           &    0.0377                        & \textbf{0.936}     & 0.0242&  \textbf{0.902}           &  0.0547           \\\bottomrule

\end{tabular}
}
\caption{Performance (average node idleness, F-score) for each algorithm tested at 0, 5, 10 and 20\% noise in measurement, averaged across 20 simulation runs. Best two performing algorithms in each category are highlighted in bold text.}
\label{tab:results-conv-fscore}
\end{table*}

\section{Conclusion and Future Work}\label{sec:discussion}

Multi-robot patrolling (MRP) is a well-studied problem with several effective control algorithms for a user to choose from. However, when there is anomaly detection noise, there is considerable untapped potential for a multi-robot system to help a user to prioritize their security inspections. This is because robots can share their detection information and form a collective consensus about which locations are most likely to be worth checking -- and such locations could be the first stops on a regular human security patrol, for example. Here, we found that MRP algorithms controlling robots with local, pairwise belief exchange, resulted in systems that tended to converge to consensus more quickly when there was higher connectivity in their emergent communications network -- that is, when they made robots physically pass by each other more often. However, for some algorithms, easy consensus formation also led to high numbers of false positive results. We developed a multi-robot F-score to obtain an overall view of system accuracy for each algorithm and noise level. We found that some of the leading MRP algorithms with respect to idleness minimization \cite{portugal_cooperative_2016}, which also had intermediate connectivity, also had good F-scores, with SEBS (State Exchange Bayesian Strategy, \cite{portugal_distributed_2013}) and CBLS (Concurrent Bayesian Learning Strategy, \cite{portugal_cooperative_2016}) having good all-round performance. Their coordination of robots to enhance efficient patrolling also has the advantage of reducing the mixing of robots, and hence constrains somewhat their local communication.

The algorithms and their emergent consensus formation behaviors have been studied here on a moderately-sized map (40 nodes, average degree 2.2). This investigation will be extended further for different maps. Given a patrol graph (map) with a different level of connectivity or larger number of nodes for the same number of robots to patrol, different interaction levels may result. This would require further investigation into the relationship between the number of robots to the number of nodes that each robot is expected to visit. Given a larger map for the same number of robots it could be anticipated that there will be fewer interactions due to the larger spatial separation between the robots (i.e. lower robot density). This may result in a slower consensus, or no consensus at all. Patrol graph connectivity (e.g. average degree) is also likely to be a relevant factor in emergent communication network connectivity.

We have examined the performance of unmodified MRP algorithms, where robot behavior does not change upon initial anomaly detection. A natural next step would be to add in a behavioral response such as a re-inspection or communication to a nearest neighbor to invite them to also visit that location. With such interaction, the spatial distribution of robots could itself indicate anomaly distribution \cite{Hunt2020,Hunt2022}. Given the good performance of algorithms resulting in \textit{intermediate} network connectivity \cite{talamali_when_2021}, this could also be used as a control input \cite{zavlanos_graph-theoretic_2011, poonawala_collision-free_2015}, with global measurement or local connectivity estimation in a decentralized implementation \cite{yang_decentralized_2008, zavlanos_distributed_2008, zavlanos_hybrid_2009}.  We will also explore further scenarios in terms  of anomaly regularity, given that real-world adversary activity is likely to be transient or intermittent. In such a case, it may well be preferable to err on the side of false positives, and prefer higher communication connectivity. Our results should help inform understanding of the trade-offs involved in using multi-robot systems to aid the perception of difficult-to-detect environmental features.  

\section*{ACKNOWLEDGEMENTS}
ZRM is supported by a University of Bristol PhD Scholarship. ERH is supported by the Royal Academy of Engineering under the Research Fellowship programme. Simulations were performed on University of Bristol Self-Service Cloud.

\bibliographystyle{ACM-Reference-Format}
\bibliography{references_date_fix_EH,refs_extra} 

\end{document}